\pdfoutput=1

\def\year{2016}\relax

\documentclass[letterpaper]{article}
\usepackage{aaai16}
\usepackage{times}
\usepackage{helvet}
\usepackage{courier}
\usepackage{graphicx}
\usepackage{amssymb}
\usepackage{amsmath}
\usepackage{fancyhdr}

\fancypagestyle{empty}{%
\fancyhead[L]{}
\fancyhead[C]{\vspace{-50mm} Presented at AAAI-16 conference, February 12-–17, 2016, Phoenix, Arizona USA}
\fancyhead[R]{}
}

\newcommand{\state}{y}
\newcommand{\hidden}{h}
\newcommand{\weights}{W}
\newcommand{\belief}{\mathcal{B}}
\newcommand{\observation}{x}
\newcommand{\prediction}{p}

\usepackage{color}
\definecolor{CommentRed}{rgb}{0.7,0,0}
\definecolor{CommentBlue}{rgb}{0,0,0.7}
\definecolor{CommentGreen}{rgb}{0,0.7,0}

\begin{document}

%
\title{Deep Tracking: Seeing Beyond Seeing Using Recurrent Neural Networks}
\author{
Peter Ondr\'{u}\v{s}ka \and Ingmar Posner\\
Mobile Robotics Group, University of Oxford, United Kingdom\\ \{ondruska, ingmar\}@robots.ox.ac.uk
}

\maketitle

\begin{abstract}
\begin{quote}
This paper presents to the best of our knowledge the first end-to-end object tracking approach which directly maps from raw sensor input to object tracks in sensor space without requiring any feature engineering or system identification in the form of plant or sensor models. Specifically, our system accepts a stream of raw sensor data at one end and, in real-time, produces an estimate of the entire environment state at the output including even occluded objects. We achieve this by framing the problem as a deep learning task and exploit sequence models in the form of recurrent neural networks to learn a mapping from sensor measurements to object tracks. In particular, we propose a learning method based on a form of input dropout which allows learning in an \emph{unsupervised} manner, only based on raw, occluded sensor data without access to ground-truth annotations. We demonstrate our approach using a synthetic dataset designed to mimic the task of tracking objects in 2D laser data -- as commonly encountered in robotics applications -- and show that it learns to track many dynamic objects despite occlusions and the presence of sensor noise.
\end{quote}
\end{abstract}

\section{Introduction}

As we demand more from our robots the need arises for them to operate in increasingly complex, dynamic environments where scenes -- and objects of interest -- are often only partially observable. However, successful decision making typically requires complete situational awareness. Commonly this problem is tackled by a processing pipeline which uses separate stages of object detection and tracking, both of which require considerable hand-engineering. Classical approaches to object tracking in highly dynamic environments require the specification of plant and observation models as well as robust data association.

\begin{figure}[t]
\begin{center}
\includegraphics[width=80mm]{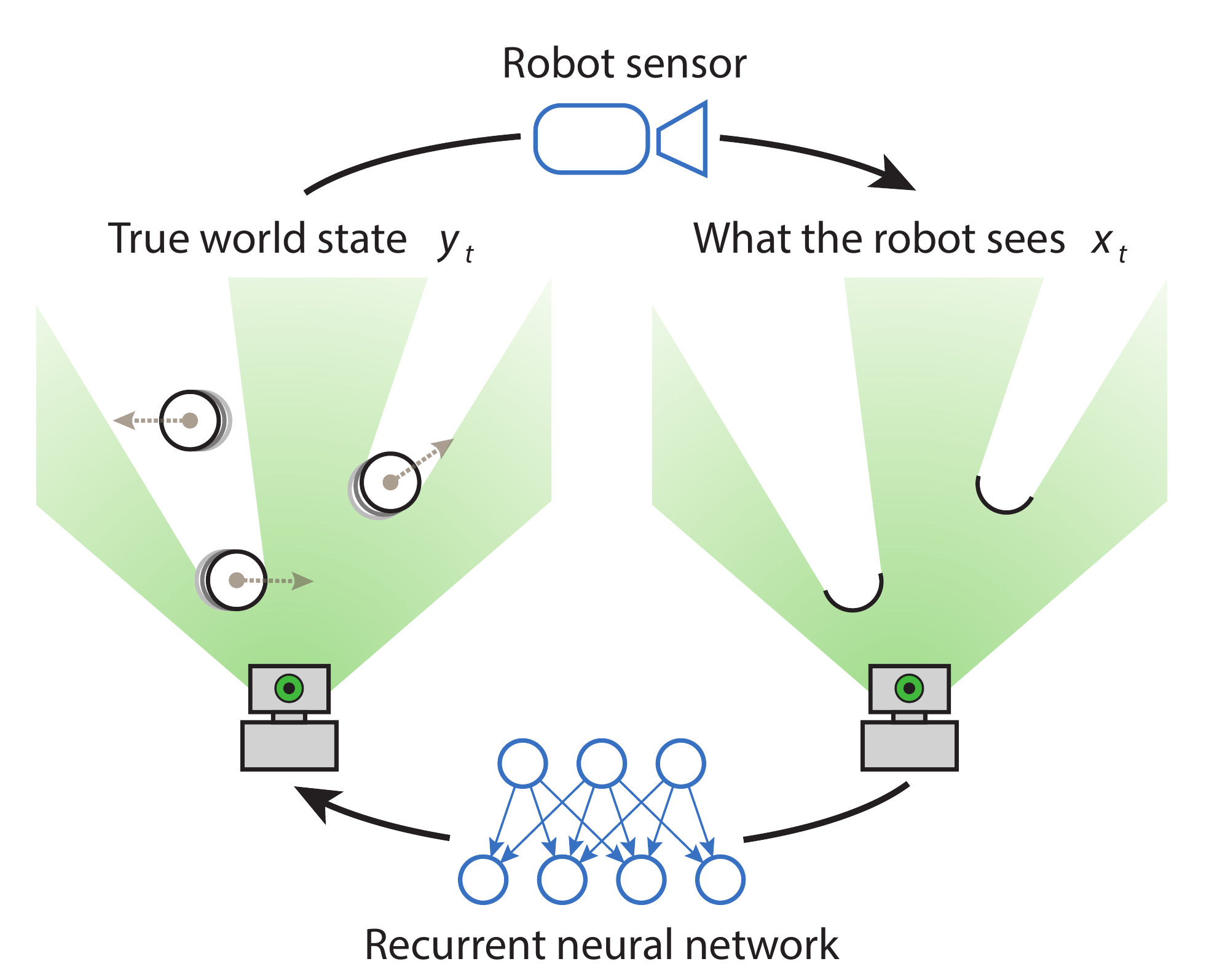}
\caption{Often a robot's sensors provide only partial observations of the surrounding environment. In this work we leverage a recurrent neural network to effectively reveal occluded parts of a scene by learning to track objects from raw sensor data -- thereby effectively reversing the sensing process.}
\label{fig:example}
\end{center}
\end{figure}

In recent years neural networks and deep learning approaches have revolutionised how we think about classification and detection in a number of domains \cite{krizhevsky2012imagenet,szegedy2013deep}. The often unreasonable effectiveness of such approaches is commonly attributed to both an ability to learn relevant representations directly from raw data as well as a vastly increased capacity for function approximation afforded by the depth of the networks \cite{bengio2009learning}.

In this work we propose a framework for \emph{deep tracking}, which effectively provides an off-the-shelf solution for learning the dynamics of complex environments directly from raw sensor data and mapping it to an intuitive representation of a complete and unoccluded scene around the robot as illustrated in Figure~\ref{fig:example}.

To achieve this we leverage Recurrent Neural Networks (RNNs), which were recently demonstrated to be able to effectively handle sequence data \cite{graves2013generating,sutskever2009recurrent}. In particular we consider a complete yet uninterpretable hidden state of the world and then train the network end-to-end to update its belief of this state using a sequence of partial observations and map it back into an interpretable unoccluded scene. This gives the network a freedom to optimally define the content of the hidden state and the considered operations without the need of any hand-engineering.

We demonstrate that such a system can be trained in an entirely \emph{unsupervised} manner based only on raw sensor data and without the need for supervisor annotation. To the best of our knowledge this is the first time such a system has been demonstrated and we believe our work will provide a new paradigm for an end-to-end tracking. In particular, our contributions can be summarized as follows:
\begin{itemize}
\item A novel, end-to-end way of filtering partial observations of dynamic scenes using recurrent neural networks to provide a full, unoccluded scene estimation. 
\item A new method of unsupervised training based on a novel form of dropout encouraging the network to correctly predict objects in unobserved spaces.
\item A compelling demonstration and attached video \footnote{Video is available at: \text{https://youtu.be/cdeWCpfUGWc}} of the method in a synthesised scenario of a robot tracking the state of the surrounding environment populated by many moving objects. The network directly accepts unprocessed raw sensor inputs and learns to predict positions of all objects in real time even through complete sensor occlusions and noise \footnote{The source code of our experiments is available at:\\ \text{http://mrg.robots.ox.ac.uk/mrg\_people/peter-ondruska/}}.
\end{itemize}


\section{Deep Tracking}
Formally, our goal is to predict the current, un-occluded scene around the robot given the sequence of sensor observations, i.e. to model $P(\state_t | \observation_{1:t})$ where $\state_t \in \mathbb{R}^N$ is a discrete-time stochastic process with unknown dynamics modelling the scene around the robot containing other static and dynamic objects and $\observation_t \subseteq \state_t$ are the sensor measurements of directly visible scene parts. We use encoding $\observation_t=\{v_t,r_t\}$ where $\{v^i_t, z^i_t\}=\{1,\state^i_t\}$ if $i$-th element of $\state_t$ is observed and $\{0,0\}$ otherwise. Moreover, as the scene changes the robot can observe different parts of the scene at different times.

In general, this problem cannot be solved effectively if the process $\state_t$ is purely random and unpredictable, as in such a case there is no way to determine the unobserved elements of $\state_t$. In practice, however, the sequences $\state_t$ and $\observation_t$ exhibit structural and temporal regularities which can be exploited for effective prediction. For example, objects tend to follow certain motion patterns and have certain appearance. This allows the behaviour of the objects to be estimated when they are visible, such that this knowledge can be used later for prediction when they become occluded.

Deep tracking leverages recurrent neural networks to effectively model $P(\state_{t}|\observation_{1:t})$. In the remainder of this section we first describe the probabilistic model assumed to underly the sensing process before detailing how RNNs can be used for tracking.

\begin{figure}[t]
\begin{center}
\includegraphics[width=75mm]{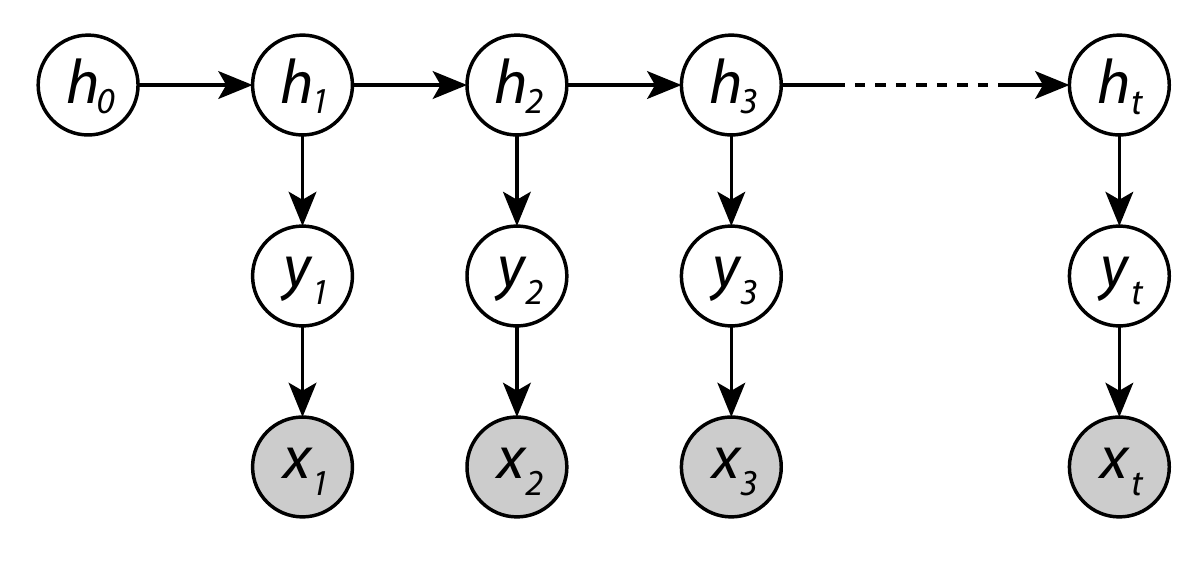}
\caption{The assumed graphical model of the generative process. World dynamics are modelled by the hidden Markov process $\hidden_t$ with appearance $\state_t$ which is partially observed by sensor measurements $\observation_t$.}
\label{fig:graph}
\end{center}
\end{figure}

\subsection{The Model}
Like many approaches to object tracking, our model is inspired by Bayesian filtering \cite{chen2003bayesian}. Note, however, the process $\state_t$ does not satisfy the Markov property: Even knowledge of $\state_t$ at any particular time $t$ provides only partial information about the state of the world such as object positions, but does not contain all necessary information such as their speed or acceleration required for prediction. The latter can be obtained only by relating subsequent measurements. Methods such as the Hidden Markov Model \cite{rabiner1989tutorial} are therefore not directly applicable.

To handle this problem we assume the generative model in Figure~\ref{fig:graph} such that alongside $\state_t$ there exists another underlying Markov process, $\hidden_t$, which completely captures the state of the world with the joint probability density
\begin{equation}
P(\state_{1:N}, \observation_{1:N}, \hidden_{1:N}) = \prod^N_{t=1} P(\observation_t | \state_t) P(\state_t | \hidden_t) P(\hidden_t | \hidden_{t-1}),
\label{eq:joint}
\end{equation}
where
\begin{itemize}
\item $P(\hidden_t | \hidden_{t-1})$ denotes the hidden state transition probability capturing the dynamics of the world;
\item $P(\state_t | \hidden_t)$ is modelling the instantaneous \emph{unoccluded} sensor space;
\item and $P(\observation_t | \state_t)$ describes the actual sensing process.
\end{itemize}
The task of estimating $P(\state_t | \observation_{1:t})$ can now be framed in the context of recursive Bayesian estimation \cite{bergman1999recursive} of the \emph{belief} $\belief_t$ which, at any point in time $t$, corresponds to a distribution $Bel(\hidden_t) = P(\hidden_t | \observation_{1:t})$. This belief can be computed recursively for the considered model as
\begin{eqnarray}
Bel^-(\hidden_t) &=& \int_{\hidden_{t-1}} P(\hidden_t | \hidden_{t-1}) Bel(\hidden_{t-1})\label{eq:BelPred}\\
Bel(\hidden_t) &\propto& \int_{\state_t} P(\observation_t | \state_t) P(\state_t | \hidden_t)
Bel^-(\hidden_t) \label{eq:BelUpdate}
\end{eqnarray}
where $Bel^-(\hidden_t)$ denotes \emph{belief prediction} one time step into the future and $Bel(\hidden_t)$ denotes the \emph{corrected} belief after the latest measurement has become available.
$P(\state_t|\observation_{1:t})$ in turn can then be expressed simply using this belief,
\begin{eqnarray}
P(\state_t | \observation_{1:t}) &=& \int_{\hidden_t} P(\state_t|\hidden_t) Bel(\hidden_t). \label{eq:emission}
\end{eqnarray}

\subsection{Filtering Using a Recurrent Neural Network}

\begin{figure}[t]
\begin{center}
\includegraphics[width=80mm]{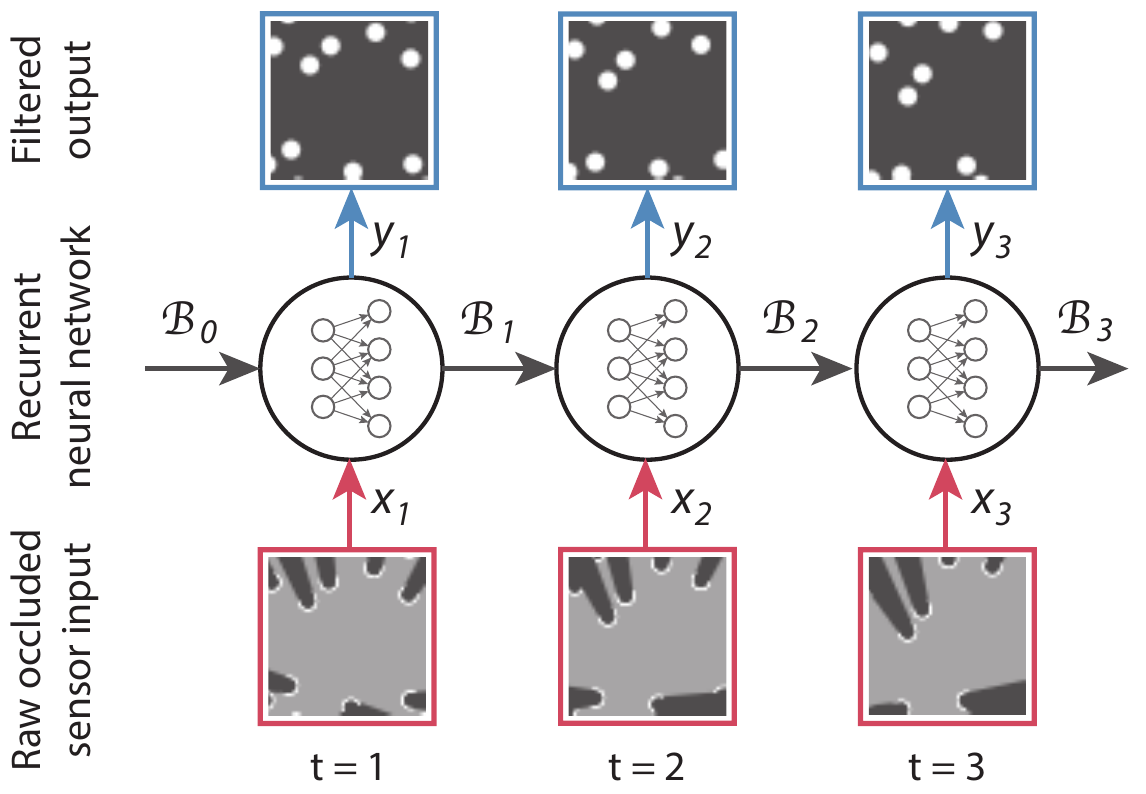}
\caption{An illustration of the filtering process using a recurrent neural network.}
\label{fig:sequence}
\end{center}
\end{figure}

Computation of $P(\state_t | \observation_{1:t})$ is carried easily by iteratively updating the current belief $\belief_t$
\begin{equation}
\belief_t = F(\belief_{t-1}, \observation_t)
\end{equation}
defined by Equations \ref{eq:BelPred},\ref{eq:BelUpdate} and simultaneously providing prediction
\begin{equation}
P(\state_t | \observation_{1:t}) = P(\state_t | \belief_t)
\end{equation}
defined by equation \ref{eq:emission}.
Moreover, instead of $\state_t$ the same method can be used to predict any state in the future $P(\state_{t+n}|\observation_{1:t})$ by providing \emph{empty} observations of the form $\observation_{(t+1):(t+n)} = \varnothing$.

For realistic applications this approach is, however, limited by the need for the suitable belief state representation as well as the explicit knowledge of the distributions in Equation \ref{eq:joint} modelling the generative process. 

The key idea of our solution is to avoid the need to specify this knowledge and instead use a highly expressive neural networks, governed by weights $\weights_F$ and $\weights_P$, to model both $F(\belief_{t-1},\observation_t)$ and $P(\state_t|\belief_t)$ allowing to learn their function directly from the data. Specifically, we assume $\belief_t$ can be approximated and represented as a vector, $\belief_t \in \mathbb{R}^M$. As illustrated in Figure \ref{fig:sequence}, the function $F$ is then simply a recurrent mapping from $\mathbb{R}^M \times \mathbb{R}^{2N} \rightarrow \mathbb{R}^M$ corresponding to an implementation by a Feed-Forward Recurrent Neural Network \cite{medsker2001recurrent} where the $\belief_t$ acts as the network's memory passed from one time step to the next.

Importantly, we do not impose any restrictions on the content or function of this hidden representation. Provided both networks for $F(\belief_t,\observation_t)$ and $P(\state_t | \belief_t)$ are differentiable they can be trained together end-to-end as a single recurrent network with a cost function directly corresponding to the data likelihood provided in Equation~\ref{eq:emission}. This allows the neural networks to adapt to each other and to learn an optimal hidden state representation for $\belief_t$ together with procedures for belief updates using partial observations $\observation_t$ and to use this information to decode the final scene state $\state_t$. Training of the network is detailed in the next section while the concrete architecture used for our experiment is discussed as a part of the Experimental Results.

\section{Training}
The network can be trained both in a supervised mode, i.e. with both known $\state_{1:N},\observation_{1:N}$ as well as in an unsupervised mode where only $\observation_{1:N}$ is known. A common method of supervised training is to minimize the negative log-likelihood of the true scene state given the sensor input
\begin{equation}
\mathcal{L} = - \sum^N_{t=1} \log P(\state_t|\observation_{1:t})
\label{eq:objective}
\end{equation}
This objective can be optimised by the gradient descent with partial derivatives given by backpropagation through time \cite{rumelhart1985learning}
\begin{eqnarray}
\frac{\partial \mathcal{L} }{ \partial \belief_{t} } &=& \frac{ \partial F(\belief_{t}, \observation_{t+1} ) }{ \partial \belief_t } \cdot \frac{\partial \mathcal{L}}{ \partial \belief_{t+1} } - \frac{ \partial \log P(\state_t | \belief_t) }{ \partial \belief_t }\\
\frac{ \partial \mathcal{L}}{ \partial \weights_F } &=& \sum^N_{t=1} \frac{\partial F(\belief_{t-1}, \observation_t)}{\partial \weights_F} \cdot \frac{\partial \mathcal{L} }{ \partial \belief_t }\\
\frac{ \partial \mathcal{L} }{ \partial \weights_P } &=& - \sum^N_{t=1} \frac{ \partial \log P(\state_t | \belief_t) }{ \partial \weights_P }.
\end{eqnarray}

However, a major disadvantage of the supervised training is the requirement of the ground-truth scene state $\state_t$ containing parts not visible by the robot sensor. One way of obtaining this information is to place additional sensors in the environment which jointly provide information about every part of the scene. Such an approach, however, can be impractical, costly or even impossible. In the following section we describe a training method which only requires raw, occluded sensor observations alone. This dramatically increases the practicality of the method, as the only necessary learning information is a long enough record of $\observation_{1:N}$, as obtained by the sensor in the field.

\subsection{Unsupervised Training}
The aim of unsupervised training is to learn $F(\belief_{t-1},\observation_t)$ and $P(\state_t | \belief_t)$ using only $\observation_{1:t}$. This might seem impossible without knowledge of $\state_t$ as there would be no way to correct the network predictions. However, \emph{some} values of $\state_t$ as $\observation_t \subseteq \state_t$ are known in the directly observed sensor measurements. A naive approach would be to train the network using the objective from Equation \ref{eq:objective} of predicting the known values of $\state_t$ and not back-propagating from the unobserved ones. Such an approach would fail, however, as the network would only learn to copy the observed elements of $\observation_t$ to the output and never predict objects in occlusion.

Instead, and crucially, we propose to train the network not to predict the current state $P(\state_t | \observation_{1:t})$ but a state in the future $P(\state_{t+n} | \observation_{1:t})$. In order for the network to correctly predict the visible part of the object after several time steps the network must learn how to represent and simulate object dynamics during the time when the object is not seen using $F$ and then convert this information to predict $P(\state_{t+n} | \belief_{t+n})$.

This form of learning can be implemented easily by considering the cost function from Equation \ref{eq:objective} of predicting visible elements of $\state_{t:t+n}$ and \emph{dropping-out} all $\observation_{t:t+n}$ by setting them to $0$ marking the input unobserved. This is similar to standard node dropout \cite{srivastava2014dropout} to prevent network from overfitting. Here, we are dropping out the observations both spatially across the entire scene and temporarily across multiple time steps in order to prevent the network to overfit to the task of simply copying over the elements. This forces the network to learn the correct patterns.

An interesting observation demonstrated in the next section is that even though the network is in theory \emph{trained to predict only the future observations} $\observation_{t+n}$, in order to achieve this the network learns to predict $\state_{t+n}$. We hypothesise this is due to the tendency of the regularized network to find the simplest explanation for observations where it is easier to model $P(\state_t | \belief_t)$ than $P(\observation_t | \belief_t)~=~\int_{\state_t}P(\observation_t|\state_t)P(\state_t|\belief_t)$.

\section{Experimental Results}

In this section we demonstrate the effectiveness of our approach in estimating the full scene state in a simulated scenario representing a robot equipped with a 2D laser scanner surrounded by many dynamic objects. Our inspiration here is a situation of a robot in the middle of a busy street surrounded by a crowd of moving people. Objects are modelled as circles independently moving with constant velocity in a random direction but never colliding with each other or with the robot. The number of objects present in the scene is varied over time between two and 12 as new objects randomly appear and later disappear at a distance from the robot.

\subsection{Sensor Input}
At each time step we simulated the sensor input as commonly provided by a planar laser scanner. We divided the scene $\state_t$ around the robot using a 2D grid of $50{\times}50$ pixels with the robot placed towards the bottom centre. To model sensor observations $\observation_t$ we ray-traced the visibility of each pixel and assigned values $\observation^i_t = \{v_t^i, r_t^i\}$ indicating pixel visibility (by the robot) and presence of an obstacle if the pixel is visible. An example of the input $\state_t$ and corresponding observation $\observation_t$ is depicted in Figure \ref{fig:sequence}. Note that, as in the case of a real sensor, only the surface of the object as seen from particular viewpoint is ever observed. This input was then fed to the network as a stream of $50{\times}50$ 2-channel binary images for filtering.

\subsection{Neural Network}

\begin{figure}[t]
\begin{center}
\includegraphics[width=80mm]{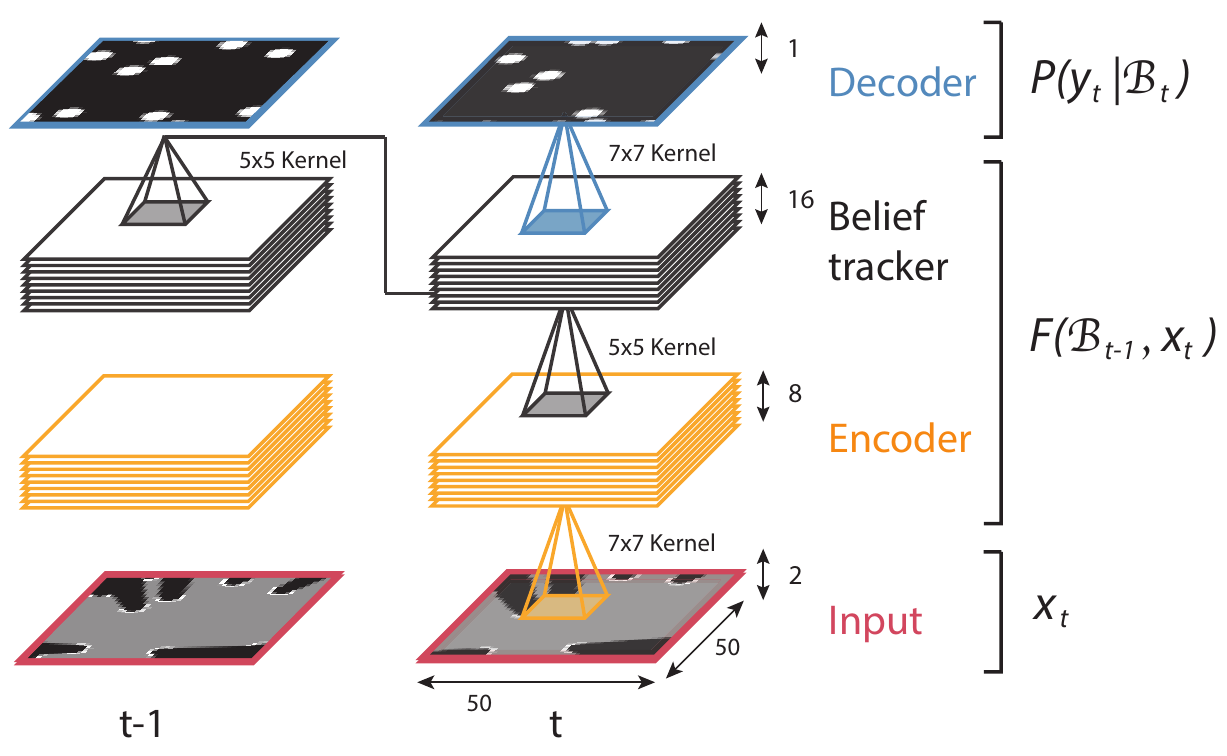}
\caption{The 4-layer recurrent network architecture used for the experiment. Directly visible objects are detected in \emph{Encoder} and fed into \emph{Belief tracker} to update belief $\belief_t$ used for scene deocclusion in \emph{Decoder}.}
\label{fig:network}
\end{center}
\vspace{-2mm}
\end{figure}

To jointly model $F(\belief_{t-1}, \observation_t)$ and $P(\state_t | \belief_t)$ we used a small feed-forward recurrent network as illustrated in Figure \ref{fig:network}. This architecture has four layers and uses convolutional operations followed by a sigmoid nonlinearity as the basic information processing step at every layer. The network has in total 11k parameters, and its hyperparameters, such as number of channels in each layer and size of the kernels, were set by cross-validation. The belief state $\belief_t$ is represented by the 3rd layer of size $50{\times}50{\times}16$ which is kept between time steps.

The mapping $F(\belief_{t-1}, \observation_t)$ is composed of the stage of input-preprocessing (the \emph{Encoder}) followed by a stage of hidden state propagation (the \emph{Belief tracker}). The aim of the \emph{Encoder} is to analyse the sensor measurements, to perform operations such as detecting objects directly visible to the sensor and to convert this information into a $50{\times}50{\times}8$ embedding as input to the \emph{Belief tracker}. This information is concatenated with the previous belief state $\belief_{t-1}$ and in \emph{Belief tracker} combined into a new belief state $\belief_t$.

Finally, $P(\state_t | \belief_t)$ is modelled by a Bernoulli distribution by interpreting the network final layer output as a \emph{probabilistic occupancy grid} $\prediction_t$ of the corresponding pixels being part of an object giving rise to the total joint probability
\begin{equation}
P(\state_t | \prediction_t) = \prod_i {(\prediction^i_t)} ^ {\state^i_t} {(1-\prediction^i_t)} ^ {(1- \state^i_t)}
\end{equation}
with its logarithm corresponding to the binary cross-entropy. One pass through the network takes 10ms on a standard laptop drawing the method suitable for real-time data filtering.

\subsection{Training}

\begin{figure*}[t]
\begin{center}
\includegraphics[width=177mm]{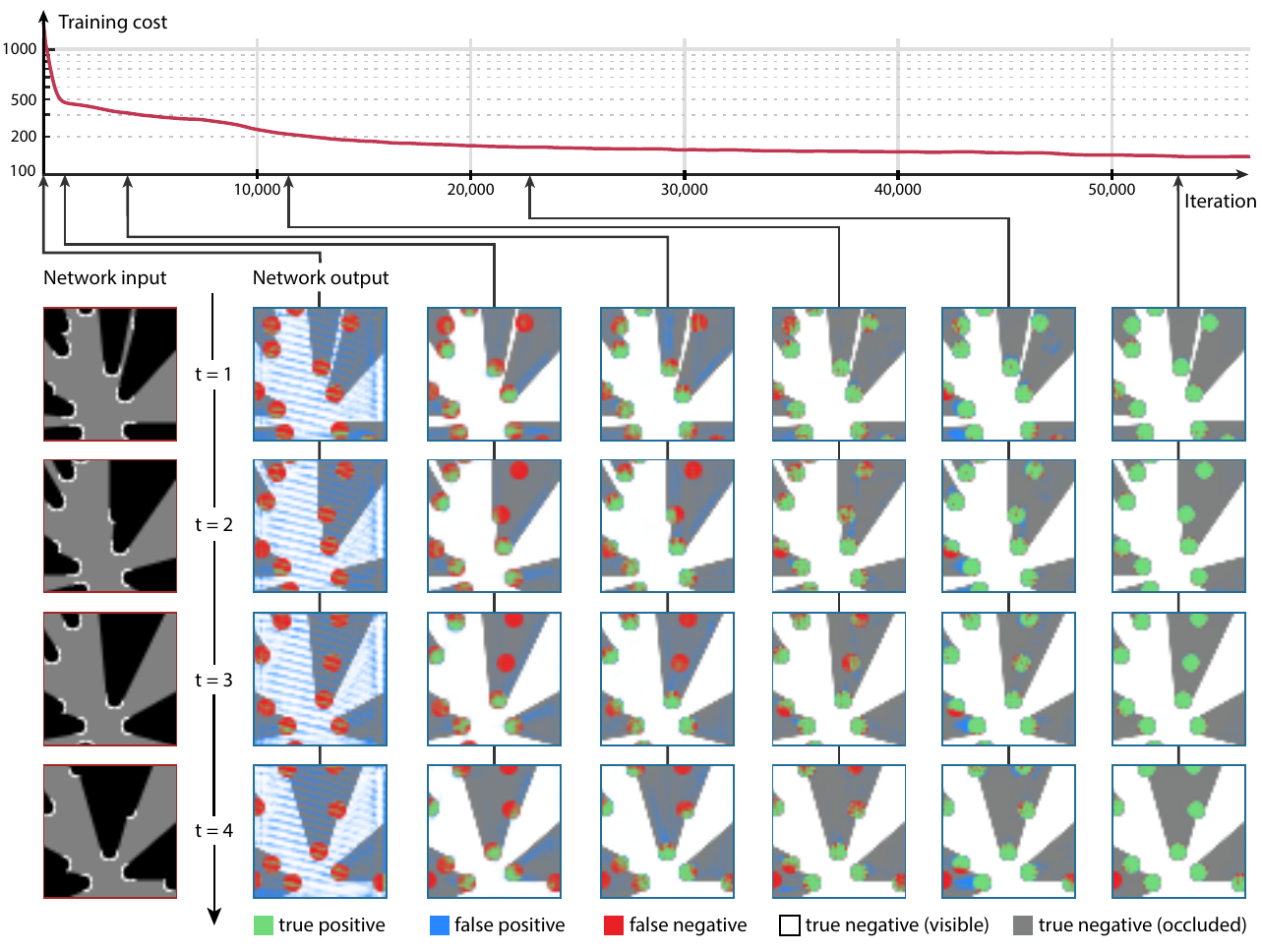}
\caption{The unsupervised training progress. From left to right, the network first learns to complete objects and then it learns to track the objects through occlusions. Some situations can not be predicted correctly such as the object in occlusion on the fourth frame not seen before. Note, the ground-truth for comparison was used only for evaluation, but not during the training.}
\label{fig:training}
\end{center}
\vspace{-3mm}
\end{figure*}

\begin{figure*}[t]
\begin{center}
\includegraphics[width=180mm]{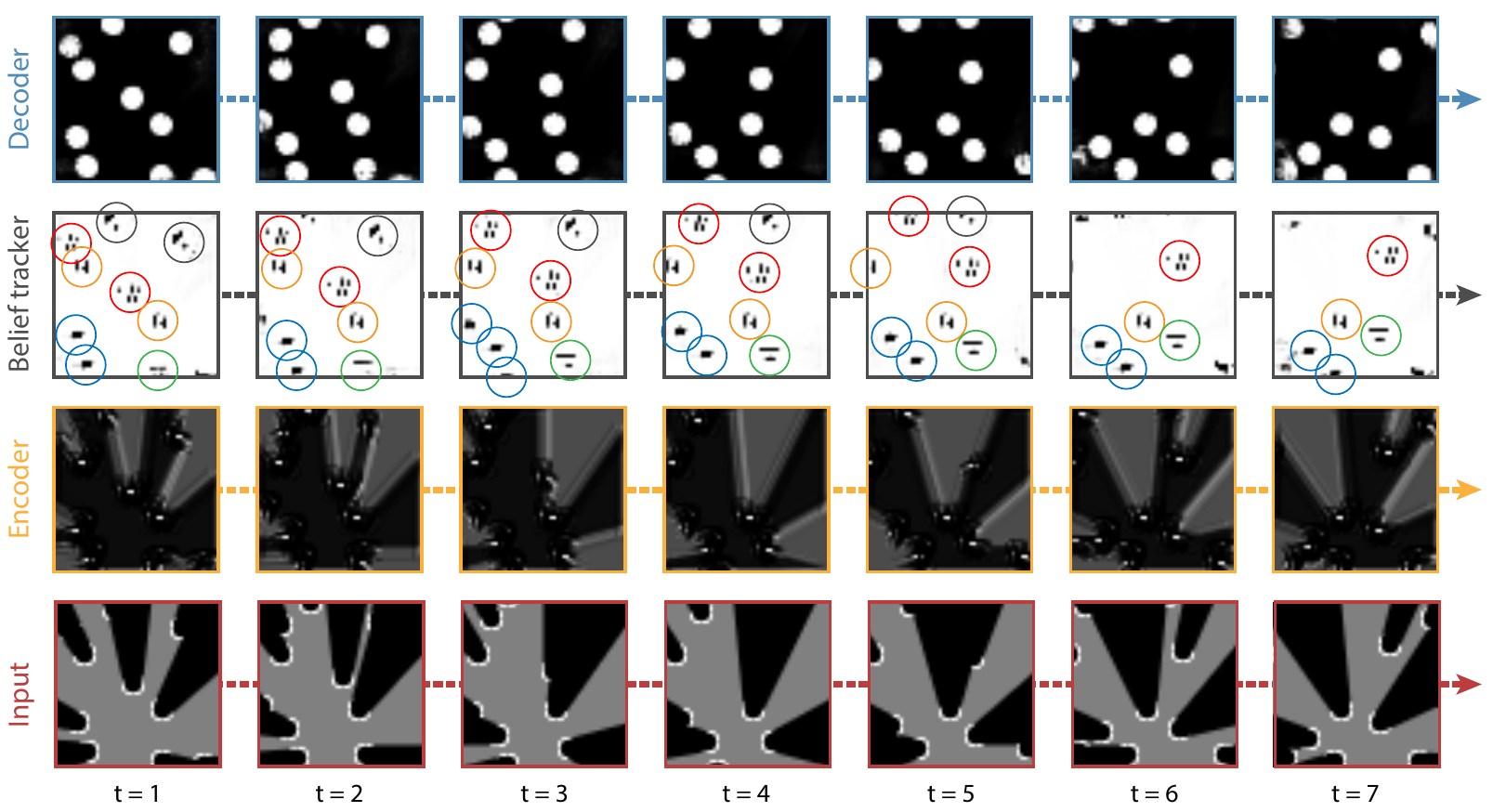}
\caption{Examples of the filter activations from different parts of the network (one filter per layer). The \emph{Encoder} adapts to produce a spike at the position of directly visible objects. Interesting is the structure of the \emph{belief} $\belief_t$. As highlighted by circles in 2nd row, it adapted to assign different activation patterns to objects having different motion patterns (best viewed in colour). }
\label{fig:analysis}
\end{center}
\vspace{-3mm}
\end{figure*}

We generated 10,000 sequences of length 200 time steps and trained the network for a total of 50,000 iterations using stochastic gradient descent with learning rate 0.9. The initial belief $\belief_t$ was modelled as a free parameter being jointly optimised with the network weights $W_F,W_P$.

Both supervised (i.e. when the groundtruth of $\state_t$ is known) and unsupervised training were attempted with almost identical results. Figure~\ref{fig:training} illustrates the unsupervised training progress. Initially, the network produces random output, then it gradually learns to predict the correct shape of the visible objects and finally it learns to track their position even through complete occlusion. As illustrated in the attached video, the fully trained network is able to confidently and correctly start tracking an object immediately after it has seen even only a part of it and is then able to confidently keep predicting its position even through long periods of occlusion.

Finally, Figure~\ref{fig:analysis} shows example activations of layers from different parts of the network. It can be seen that the \emph{Encoder} module learned to produce in one of its layers a spike at the center of visible objects. Particularly interesting is the learned structure of the \textit{belief state}. Here, the network learned to represent hypotheses of objects having different motion patterns with different patterns of unit activations and to track this pattern from frame to frame. None of this behaviour is hard-coded and it is a pure result of the network adaptation to the task.

Additionally we performed experiments changing the shape of objects from circles to squares and also simulating sensor noise on 1\% of all pixel observations\footnote{See the attached video.}. In all cases the network correctly learned and predicted the world state, suggesting the approach and the learning procedure is able to perform well in a variety of scenarios. We expect more complex cases will require networks having more intermediate layers or using different units such as LSTM \cite{hochreiter1997long}, however.

\section{Related Works}
Our work concerns modelling partially-observable stochastic processes with a particular focus on the application of tracking objects. This problem is commonly solved by Bayesian filtering which gives rise to tracking methods for a variety of domains \cite{yilmaz2006object}. Commonly, in these approaches the state representation is designed by hand and the prediction and correction operations are made tractable under a number of assumptions on model distributions or via sampling-based methods. The Kalman filter \cite{kalman1960new}, for example, assumes a multivariate normal distribution to represent the belief over the latent state leading to the well-known prediction/update equations but limiting its expressive power. In contrast, the Particle filter \cite{thrun2005probabilistic} foregoes any assumptions on belief distributions and employs a sample-based approach. In addition, where multiple objects are to be tracked, correct data association is crucial for the tracking process to succeed.

In contrast to using such hand-designed pipelines we provide a novel end-to-end trainable solution allowing the robot to automatically learn an appropriate belief state representation as well as the corresponding predict and update operations for an environment containing multiple objects with different appearance and potentially very complex behaviour. While we are not the first to leverage the expressive power of neural networks in the context of tracking, prior art in this domain primarily concerned only the detection part of the tracking pipeline \cite{fan2010human}, \cite{janen2010multi}.

A full neural network pipeline requires the ability to learn and simulate the dynamics of the underlying system state in the absence of measurements such as when the tracked object becomes occluded. Modelling complex dynamic scenes however requires modelling distributions of exponentially-large state representations such as real-valued vectors having thousands of elements. A pivotal work to model high-dimensional sequences was based on using Temporal Restricted Boltzman Machines \cite{sutskever2007learning,sutskever2009recurrent,mittelman2014structured}. This generative model allows modelling the joint distribution of $P(\state, \observation)$. The downside of using an RBM, however, is the need of sampling, making the inference and learning computationally expensive. In our work, we assume an underlying generative model but, in the spirit of Conditional Random Fields \cite{lafferty2001conditional} we directly model only $P(\state | \observation)$. Instead of RBM our architecture features Feed-forward Recurrent Neural Networks \cite{medsker2001recurrent,graves2013generating} making the inference and weight gradients computation exact using a standard back-propagation learning procedure \cite{rumelhart1988learning}.
Moreover, unlike undirected models, a feed-forward network allows the freedom to straightforwardly apply a wide variety of network architectures such as fully-connected layers and LSTM \cite{hochreiter1997long} to design the network. Our used architecture is similar to recent Encoder-Recurrent-Decoder \cite{fragkiadaki2015recurrent} and similarly to \cite{shi2015convolutional} which features convolutions for spatial processing.

\section{Conclusions}
In this work we presented \emph{Deep Tracking}, an end-to-end approach using recurrent neural networks to map directly from raw sensor data to an interpretable yet hidden sensor space, and employ it to predict the unoccluded state of the entire scene in a simulated 2D sensing application. The method avoids any hand-crafting of plant or sensor models and instead learns the corresponding models directly from raw, occluded sensor data. The approach was demonstrated on a synthetic dataset where it achieved highly faithful reconstructions of the underlying world model.

As future work our aim is to evaluate \emph{Deep Tracking} on real data gathered by robots in a variety of situations such as pedestrianised areas or in the context of autonomous driving in the presence of other traffic participants. In both situations knowledge of the likely unoccluded scene is a pivotal requirement for robust robot decision making. We further intend to extend our approach to different modalities such as 3D point-cloud data and depth cameras.

\bibliography{main}
\bibliographystyle{aaai}

\end{document}